\documentclass[10pt, conference, compsocconf]{IEEEtran}

\newcommand{\hpcasubmissionnumber}{69}

\usepackage[normalem]{ulem}
\usepackage{psfig}
\usepackage{graphics}
\usepackage{comment}
\usepackage[normalem]{ulem}
\usepackage{url}
\usepackage{psfig}
\usepackage{graphics}
\usepackage{comment}
\usepackage{graphicx}
\usepackage[normalem]{ulem}
\usepackage{amssymb,amsmath}
\usepackage{upgreek}
\usepackage{tikz}
\usepackage{enumitem}

\newcommand{\name}{Newton }
\newcommand{\names}{Newton}

\begin{document}

\title{
\names: Gravitating Towards the Physical Limits of Crossbar Acceleration
}

\author{
\IEEEauthorblockN{Anirban Nag\IEEEauthorrefmark{1}, Ali Shafiee\IEEEauthorrefmark{1}, Rajeev Balasubramonian\IEEEauthorrefmark{1}, Vivek Srikumar\IEEEauthorrefmark{1}, Naveen Muralimanohar\IEEEauthorrefmark{2}}
\IEEEauthorblockA{\IEEEauthorrefmark{1}School of Computing, University of Utah, Salt Lake City, Utah, USA\\Email: \{anirban, shafiee, rajeev, svivek\}@cs.utah.edu}
\IEEEauthorblockA{\IEEEauthorrefmark{2}Hewlett Packard Labs, Palo Alto, California, USA\\Email: \{naveen.muralimanohar\}@hpe.com}
}

\date{}
\maketitle

\thispagestyle{empty}

\begin{abstract}
Many recent works have designed accelerators for Convolutional Neural
Networks (CNNs).  While digital accelerators have relied on near data
processing, analog accelerators have further reduced data movement by
performing in-situ computation.  Recent works
take advantage of highly parallel analog in-situ computation in
memristor crossbars to accelerate the many vector-matrix multiplication
operations in CNNs.  However, these in-situ accelerators
have two significant short-comings that we address in this work.
First, the ADCs account for a large fraction of chip power and area.  
Second, these accelerators adopt a homogeneous design
where every resource is provisioned for the worst case.
By addressing both problems, the new architecture, \names, moves closer
to achieving optimal energy-per-neuron for crossbar accelerators.

We introduce multiple new techniques that apply at different levels of
the tile hierarchy.  Two of the techniques leverage heterogeneity:
one adapts ADC precision based on the requirements of every sub-computation
(with zero impact on accuracy),
and the other designs tiles customized for convolutions or classifiers.
Two other techniques rely on divide-and-conquer numeric algorithms
to reduce computations and ADC pressure.  Finally, we
place constraints on how a workload is mapped to tiles, thus helping
reduce resource provisioning in tiles.
For a wide range of CNN dataflows and structures,
\name achieves a 77\% decrease in power, 51\% improvement in energy
efficiency, and $2.2\times$ higher throughput/area, relative to the
state-of-the-art ISAAC accelerator.

\end{abstract}

\vspace{-0.10in}
\section{Introduction}
\label{intro}
\vspace{-0.05in}
Accelerators are in vogue today, primarily because it is evident that
annual performance improvements can be sustained via specialization.
There are also many emerging applications that demand high-throughput
low-energy hardware, such as the machine learning tasks that are
becoming commonplace in enterprise servers, self-driving cars, and
mobile devices.  The last two years have seen a flurry of activity
in designing machine learning accelerators~\cite{chendu14, chenluo14, dufasthuber15, liuchen15, chili16, jizhang16, hanliu16, reaganwhatmough16,kimkung16}.
Similar to our work, most of these recent works have focused on
inference in artificial neural networks, and specifically deep convolutional
networks, that achieve state-of-the-art accuracies on challenging
image classification workloads.

While most of these recent accelerators have used digital architectures~\cite{chendu14,chenluo14},
%These accelerators typically dedicate a small fraction of chip area for
%ALUs, and store weights and neuron outputs in large SRAM/eDRAM banks.
%The different layers of a network are time-multiplexed on the ALUs,
%leveraging near-data processing to reduce data movement costs.
%
a few have leveraged analog acceleration on
memristor crossbars~\cite{shafieenag16,chili16,bojnordiipek16}.  Such
accelerators take advantage of in-situ computation to dramatically reduce
data movement costs.  Each crossbar is assigned to execute parts of
the neural network computation and programmed with the corresponding
weight values.  Input neuron values are fed to the crossbar, and by
leveraging Kirchoff's Law, the crossbar outputs the corresponding
dot product.  The neuron output undergoes analog-to-digital conversion (ADC)
before being sent to the next layer.  Multiple small-scale prototypes
of this approach have also been demonstrated~\cite{hpvid, mercedemmanuelle16}.

The design constraints for digital accelerators are very different from their
analog designs. High communication overhead and the memory bottleneck are still 
first order design constraints in digital, whereas the computation overhead
arising from analog-to-digital and digital-to-analog conversions, and balancing 
the extent of digital computation in an analog architecture are more critical in analog 
accelerators.
In this work, we show that computation is a critical problem in analog and
leverage numeric algorithms to reduce conversion overheads. Once we improve the
efficiency of computation, the next major overhead comes from communication and
storage. Towards this end, we discuss mapping techniques and buffer management
strategies to further improve analog accelerator efficiency.
%In this work, we focus on innovations to the recent ISAAC
%architecture~\cite{shafieenag16}.  While ISAAC was able to provide an
%order of magnitude improvement in throughput, relative to state-of-the-art
%digital architectures, it has a higher power density.  This is primarily
%because of the power and area overheads of ADC units.  Therefore, 
%we try to improve various aspects of the ISAAC design, with a special
%emphasis on reducing ADC and other resource requirements. 
%
%We introduce six innovations at different levels of the tile hierarchy.
%These innovations leverage heterogeneous requirements of
%different parts of the neural network computation.  They avoid over-provisioning
%ADC, HTree, and buffer resources.  And finally, they leverage numeric algorithms
%to further reduce the involvement of ADCs.

With these innovations in place, our new design, \names, moves the analog architecture
closer to the bare minimum energy required to
process one neuron. 
%It does this by reducing the overheads
%imposed by the architecture and by modern large workloads.  
We define an ideal neuron as one that keeps the weight in-place
adjacent to a digital ALU, retrieves the input from an adjacent single-row
eDRAM unit, and after performing one digital operation, writes the result to
another adjacent single-row eDRAM unit.  This energy is lower than that
for a similarly ideal analog neuron because of the ADC cost.  This ideal
neuron operation consumes 0.33~pJ.  An average DaDianNao operation consumes 
3.5~pJ because it pays a high price in data movement for inputs and weights.
ISAAC~\cite{shafieenag16} is a state-of-the-art analog design
that achieves an order of magnitude better performance than digital
accelerators such as DaDianNao.
An average ISAAC operation consumes 1.8~pJ because it pays a moderate price
in data movement for inputs (weights are in-situ) and a high price for ADC. 
An average Eyeriss~\cite{chenemer16} operation consumes 1.67~pJ because of
an improved dataflow to maximize reuse.
The innovations in \name push the analog architecture closer to the
ideal neuron by consuming 0.85~pJ per operation.
Relative to ISAAC, \name achieves a 77\% decrease in power, 51\%
decrease in energy, and $2.2\times$ increase in throughput/area.

\section{Background}
\label{back}
\vspace{-0.05in}
\subsection{Workloads}
\label{work}
\vspace{-0.05in}
We consider different CNNs presented in the ILSVRC challenge of image 
classification for the IMAGENET \cite{russakovskyolga14} dataset. The suite of benchmarks considered
in this paper is representative of the various dataflows in such image
classification networks.  For example, Alexnet is the simplest of CNNs
with a reasonable accuracy, where a few convolution layers at the start
extract features from 
the image, followed by fully connected layers that classify the image. The other networks 
were designed with a similar structure but made deeper and wider with more parameters. 
For example, MSRA Prelu-net \cite{hezhang15} has 14 more layers than Alexnet \cite{krizhevskysutskever12} and has 330 million parameters, 
which is 5.5$\times$ higher than Alexnet. On the other hand, residual nets have forward connections 
with hops, i.e., output of a layer is passed on to not only the next layer but subsequent layers. 
Even though the number of parameters in Resnets \cite{hezhang15b} are much lower, these networks are much
deeper and have a different dataflow, which changes the buffering requirements in 
accelerator pipelines.

\subsection{The Landscape of CNN Accelerators}
\label{land}
\vspace{-0.05in}
\noindent {\bf Digital Accelerators.}
The DianNao~\cite{chendu14} and DaDianNao~\cite{chenluo14} accelerators
were among the first to target deep convolutional networks at scale.  DianNao
designs the digital circuits for a basic NFU (Neural Functional Unit).
DaDianNao is a tiled architecture where each tile has an NFU and eDRAM
banks that feed synaptic weights to that NFU.  DaDianNao uses many tiles
on many chips to parallelize the processing of a single network layer.
Once that layer is processed, all the tiles then move on to processing the
next layer in parallel. 
Recent papers, e.g., Cnvlutin~\cite{albericiojudd16}, have modified
DaDianNao so the NFU does not waste time and energy processing zero-valued
inputs.  EIE~\cite{hanliu16} and Minerva~\cite{reaganwhatmough16} address 
sparsity in the weights.
%it is able
%to compress the weights by pruning away weights that are close to zero, and
%with a quantization process.  This allows a large set of network weights to be
%accommodated on a single chip; but because of the necessary sparse
%representation, the NFU loses its SIMD-ness.  Minerva~\cite{reaganwhatmough16}
%uses similar pruning and quantization, while also lowering SRAM voltages
%in a fault-aware manner.
Eyeriss~\cite{chenemer16} and ShiDianNao~\cite{dufasthuber15}
improve the NFU dataflow to maximize operand reuse.
A number of other digital
designs~\cite{kimkung16, liudu16, gaopu17} have also emerged in the past year.

\noindent {\bf Analog Accelerators.}
Two CNN accelerators introduced in the past year, ISAAC~\cite{shafieenag16}
and PRIME~\cite{chili16}, have leveraged memristor crossbars to perform
dot product operations in the analog domain.  We will focus on ISAAC
here because it out-performs PRIME in terms of throughput, accuracy,
and ability to handle signed values.  ISAAC is also able to achieve
nearly 8$\times$ and 5$\times$ higher throughput than digital accelerators
DaDianNao and Cnvlutin respectively.

\subsection{ISAAC}
\label{isaac}
\vspace{-0.05in}
\noindent {\bf Pipeline of Memristive Crossbars.}
In ISAAC, memristive crossbar arrays are used to perform analog dot-product
operations.  Neuron inputs are provided as voltages to wordlines; neuron
weights are represented by pre-programmed cell conductances; neuron outputs
are represented by the currents in each bitline.
The neuron outputs are processed by an ADC and shift-and-add circuits.
They are then sent as inputs to the next layer of neurons.
As shown in Figure~\ref{isaac}, ISAAC is a tiled
architecture; one or more tiles are dedicated to process one layer of the
neural network.  To perform inference for one input image, neuron outputs
are propagated from tile to tile until all network layers have been
processed.

\begin{comment}
In ISAAC, a matrix of
neural network weights is 
programmed as cell conductances in a crossbar, and inputs are applied as
voltages to the wordlines of the 
crossbar. The emerging bitline currents in the crossbar represent
the dot product of input voltages and cell conductances in that column;
they therefore 
the neuron output values.  These analog current values are stored in sample and hold circuits
and converted to digital 
values using high frequency Analog to Digital Converters (ADC).  Given high memristor
write latencies, a crossbar cannot be used by different network layers in a time-multiplexed
fashion. 
ISAAC is designed as a throughput architecture, where all the 
network weights are pre-programmed across different crossbars, and the dataflow is pipelined 
in order to keep all the crossbars busy in every cycle.  For example, once image 1 is processed
by network layer 1 on crossbar 1, it moves on to processing layer 2 on crossbar 2; simultaneously,
image 2 starts processing layer 1 on crossbar 1.

\end{comment}

\noindent {\bf Tiles, IMAs, Crossbars.}
An ISAAC chip consists of many tiles connected in a mesh topology 
(Figure~\ref{isaac}).  Each tile includes an eDRAM buffer that supplies
inputs to In-situ Multiply Accumulate (IMA) units.  The IMA units
consist of memristor crossbars that perform the dot-product computation,
ADCs, and shift-and-add circuits that accumulate the digitized results.
With a design space exploration, the tile is provisioned with an optimal
number of IMAs, crossbars, ADCs, etc.
Within a crossbar, a 16-bit weight is stored 2 bits per cell, across 8 columns.
A 16-bit input is supplied as voltages over 16 cycles, 1 bit per cycle,
using a trivial DAC array. The partial outputs are
shifted and added across 8 columns, and across 16 cycles to give the output
of $16b\times16b$ MAC operations.  Thus, there are two
levels of pipelining in ISAAC: (i) the intra-tile pipeline, where inputs
are read from eDRAM, processed by crossbars in 16 cycles, and aggregated,
(ii) the inter-tile pipeline, where neuron outputs are transferred
from one layer to the next.
The intra-tile pipeline has a cycle time of 100~ns, matching the latency
for a crossbar read. 
Inputs are sent to a crossbar in an IMA using an input h-tree network.
The input h-tree has sufficient bandwidth to keep all crossbars active
without bubbles. 
Each crossbar has a dedicated
ADC operating at 1.28~$GSample/s$ shared across its 128 bitlines to convert the
analog output to digital in 100~ns.  An h-tree network is then used to collect
digitized outputs from crossbars.

\begin{comment}

ISAAC is a tiled architecture, where a chip consists of a number of tiles connected in a mesh 
topology, shown in Figure~\ref{isaac}. Four such tiles share a router to send output neuron values to tiles computing 
the next layer. Each such tile consists of an eDRAM buffer that supplies 
input neurons via a bus to In-situ Multiply Accumulate (IMA) units. The IMA units consist of 
memristor crossbars that perform the dot-product computation and shift-and-add circuits
that accumulate digitized results. We carried out a design space exploration to identify that
computational efficiency (throughput/area) is optimized with 8 IMAs per 
tile and 8 crossbars per IMA.  Meanwhile, power efficiency (operations/watt) is optimized
with 16 IMAs per tile and 8 crossbars per IMA.

\end{comment}

\begin{figure}
\centering
\includegraphics[scale=0.95]{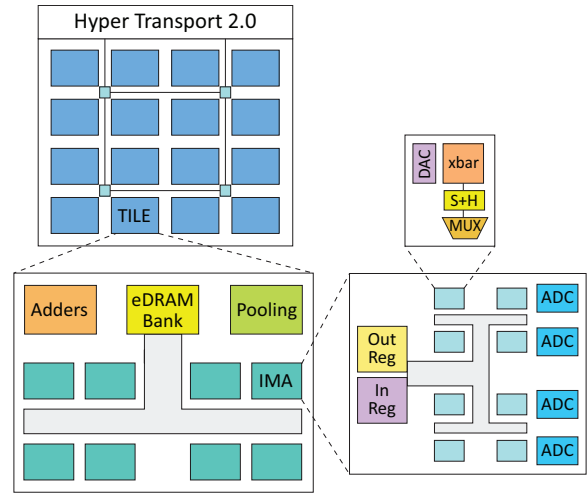}
%\vspace{-0.15in}
\caption{The ISAAC Architecture.}
\label{isaac}
\vspace{-0.15in}
\end{figure}

\vspace{-0.02in}
\noindent {\bf Crossbar Challenges.}
As with any new technology, a memristor crossbar has
unique challenges, mainly in two respects. First, mapping
a matrix onto a memristor crossbar array requires
programming (or writing) cells with the highest precision
possible. Second, real circuits deviate from ideal
operation due to parasitics such as wire resistance, device
variation, and write/read noise. All of these factors
can cause the actual output to deviate from its ideal
value.  Recent work~\cite{hustrachan16} has captured many of
these details to show the viability of prototypes~\cite{hpvid}.
The Appendix summarizes some of these details.

\section{Proposal}
%\section{The \name Architecture}
\label{proposal}

The design constraints for digital accelerators are very different from their
analog counterparts. In any digital design, the overhead of communication,
arising from the need to fetch both input feature and weights from memory, is
the major limiting factor. As a result, most optimizations focus on improving
memory bandwidth (e.g., HBM or GDDR), memory utilization (compression, zero
value elimination, etc.) and scheduling (e.g., batching) to improve
communication efficiency and 
performance. Techniques that primarily target improving digital computation
should carefully consider their impact on additional on-chip storage and
communication overheads, which can negatively affect overall efficiency.

In analog, because we do in-situ computation, only one of the operands needs to
be transferred, and this reduces the communication overhead by at least 2$\times$. 
Furthermore, the transferred value (the input vector in the form of crossbar
row voltage) is streamed across the entire crossbar (matrix values) guaranteeing
high reuse and locality. The
compute density of analog in-situ units is also better than digital accelerators. As
analog crossbars  store neural network weights, even if they are not
performing computation, they still act as on-chip storage. Whereas, digital computational
units need to have high utilization to maximize performance, otherwise their
area is better utilized for more on-chip storage.  Both these factors provide
more flexibility for analog accelerators to explore computational optimizations
at the expense of either more communication or crossbar storage.

In a digital design, the datapath size and its overhead are pre-determined. A
16-bit datapath operated with 12-bit values will achieve only marginal
reduction in overhead as pipeline buffers and wire repeaters switch every cycle.
However, as analog computation is being performed at bit
level (1 or 2 bit computations in each bitline), reducing the operand size,
say, from 16-bits to 12-bits will correspondingly reduce ADC and DAC usage,
leading to better efficiency. Note that even though an analog architecture
consists of both digital and analog computations, the overhead of analog
dominates - 61\% of the total power~\cite{shafieenag16}. 

\begin{figure}[h!]
\centering
\includegraphics[width=\columnwidth]{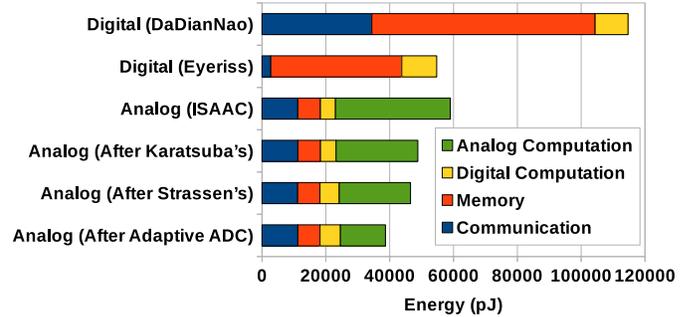}
\vspace{-0.20in}
\caption{Energy Breakdown of Vector-Matrix Multiplication in existing Digital and Analog pipelines and for the proposed optimizations}
\label{digvanabreakdown}
\vspace{-0.00in}
\end{figure}

We will first take a closer look at a simple dot-product being performed using
crossbars.  
Consider a 1$\times$128 vector being multiplied with a 128$\times$128 matrix 
(all values are 16 bits).
Figure~\ref{digvanabreakdown}  shows the  energy breakdown of the
 vector-matrix multiplication pipeline compared against  digital designs
for various architectures. To model the analog overhead, we consider 2-bit
cells, 1-bit DAC, and 16-bit values interleaved across eight crossbars. 
In a single iteration, a crossbar column is performing
a dot-product involving 128 rows, 1-bit inputs, and 2-bit cells; it
therefore produces a 9-bit result requiring a 9-bit ADC\footnote{
        Prior work (ISAAC) has shown that simple data encoding schemes can reduce the ADC resolution by
        1 bit~\cite{shafieenag16}.
}.

We must shift and add the results of eight such columns, yielding a 23-bit
result.  These results must also be shifted and added across 16 iterations,
finally yielding a 39-bit output. Finally, the scaling factor is applied to
convert the 39-bit result to a 16-bit output.
As the figure shows, communication and memory accesses are the major limiting
factor for digital architectures, whereas for analog, computation overhead,
primarily arising from ADC dominates. 

Based on these observations, we present optimizations that exploit high
compute density and flexible datapath enabled by analog to improve computation
efficiency. These optimizations are applicable to any accelerator
that uses analog in-situ crossbars as the techniques primarily target high ADC overhead. 
Once we improve the efficiency of the computation, the next major overhead
comes from communication of values.
As communication (on-chip and off-chip)
and storage overheads (SRAM or eDRAM buffers) depend on the overall accelerator
architecture, we choose the ISAAC architecture as the baseline when discussing
our optimizations.

\subsection{Reducing Computational Overhead}
\label{dandc}

\subsubsection{ Karatsuba's Divide and Conquer Multiplication Technique} 

With ADC being the major contributor to the total power, we discuss a divide and conquer 
strategy at the bit level, that reduces  pressure on ADC usage and hence ADC power. 
A classic multiplication approach for two n-bit numbers has a complexity of
$O(n^{2})$ where each bit of a number is multiplied with n-bits of the other
number, and the partial results are shifted and added to get the final
2n-bit result. 

\begin{figure}[h!]
\centering
\includegraphics[width=\columnwidth]{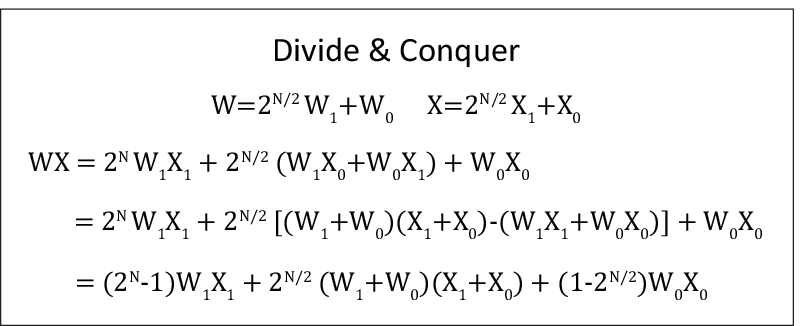}
\vspace{-0.15in}
\caption{Karatsuba's Divide \& Conquer Algorithm.}
\label{fig:karatsuba}
\vspace{-0.15in}
\end{figure}

Karatsuba's divide and conquer algorithm manages to reduce the complexity from $O(n^{2})$ to  $O(n^{1.5})$. As 
shown in Figure~\ref{fig:karatsuba}, it divides the numbers into two halves of n/2 bits, MSB bits and LSB bits, and instead of
performing four smaller n/2-bit multiplications, it calculates the result with
two n/2-bit multiplications and one (n/2 + 1)-bit multiplication.

To illustrate the benefit of this technique, consider the same example
discussed earlier using 128x128 crossbars, 2-bit cells, and 1-bit DAC.
The product of input $X$ and weight $W$ is performed
on 8 crossbars in 16 cycles (since each weight is spread across 8 cells in
8 different crossbars and the input is spread across 16 iterations).
In the example in Figure~\ref{fig:karatsuba}, $W_0X_0$ is performed
on four crossbars in 8 iterations (since we are dealing with fewer bits for
weights and inputs).  The same is true for $W_1X_1$.  A third set of
crossbars stores the weights $(W_1+W_0)$ and receives the pre-computed
inputs $(X_1+X_0)$.  This computation is spread across 5 crossbars and
9 iterations.  We see that the total amount of work has reduced by 15\%.

There are a few drawbacks as well.  A computation now takes 17 iterations
instead of 16.  The net area increases because the network  must send inputs
$X_0$ and $X_1$ in parallel, an additional crossbar is needed, the output
buffer is larger to store subproducts, and 128 1-bit full adders are
required to compute $(X_1+X_0)$.  Again, given that the ADC is the primary
bottleneck, these other overheads are relatively minor.

\subsubsection{Strassen's Algorithm}
\label{strassen}

\begin{figure}[h!]
        \centering
        \includegraphics[scale=1.0]{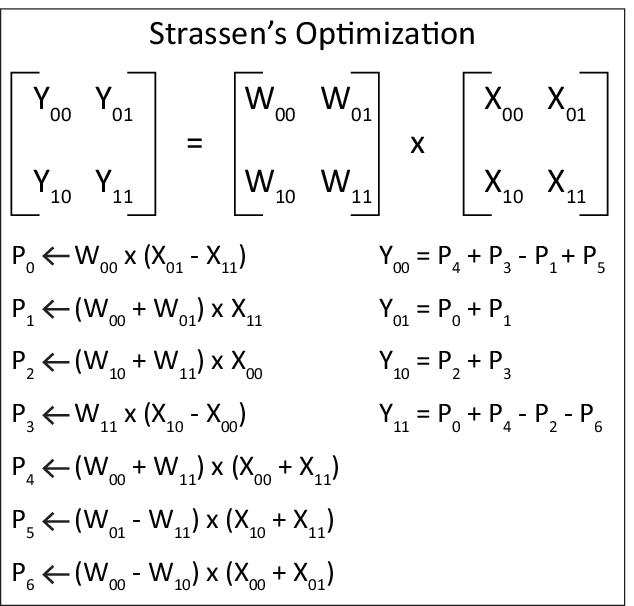}
        %\vspace{-0.1in}
        \caption{Strassen's Divide \& Conquer Algorithm for Matrix
Multiplication.}
\label{fig:strassen}
\vspace{-0.15in}
\end{figure}

A divide and conquer approach can also be applied to matrix-matrix
multiplication.  
By  partitioning each matrix $X$ and $W$ into 4 sub-matrices, we can 
express matrix-matrix multiplication in terms of multiplications
of sub-matrices.  A typical algorithm 
would require 8 sub-matrix multiplications, followed by an aggregation step.
But as shown in Figure~\ref{fig:strassen}, linear algebra manipulations
can perform the same computation with 7 sub-matrix multiplications,
with appropriate pre- and post- processing.  Similar to Karatsuba's
algorithm, this has the advantage of reducing ADC usage and power.

The above two optimizations reduce the computation
energy by 20.6\% while incurring a storage overhead of 4.3\%.
While both divide and 
conquer algorithms (Karatsuba's and Strassen's algorithms)
are highly effective for a crossbar-based architecture, they
have very little impact on other digital accelerators.  For example, 
these algorithms may impact the efficiency of the NFUs in DaDianNao,
but DaDianNao area is dominated by eDRAM banks and not NFUs.  In fact,
Strassen's algorithm can lower DaDianNao efficiency because buffering
requirements may increase.  On the 
other hand, analog computations are dominated by ADCs,
so efficient computation does noticeably impact overall efficiency.
Further, some of the pre-processing for these algorithms is performed when
installing weights on analog crossbars, but has to be performed on-the-fly
for digital accelerators.

\subsubsection{Adaptive ADCs.}
\label{heteroadc}

A simple dot-product operation on 16-bit values performed using crossbars typically
result in an output of more than 16-bits. In the example discussed earlier, using
2-bit cells in crossbars and 1-bit DACs yielded  39-bit output.  Once the scaling factor is applied, the
least significant 10 bits are dropped.  The most significant 13 bits
represent an overflow that cannot be captured in the 16-bit result, so they
are effectively used to clamp the result to a maximum value.

\begin{figure}[h!]
        \centering
        \includegraphics[scale=0.45]{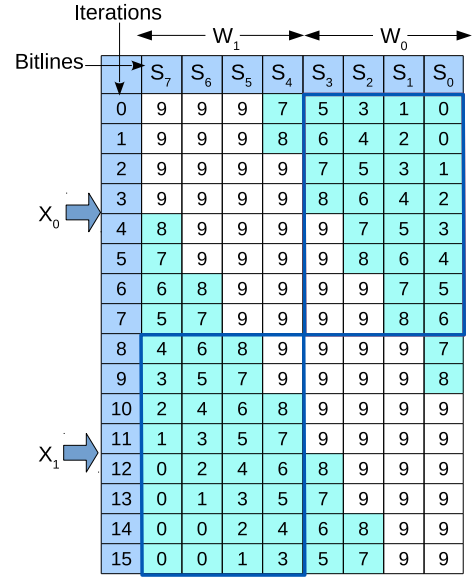}
        %\vspace{-0.1in}
        \caption{Heterogeneous ADC sampling resolution.}
        \label{adcgrid}
        \vspace{-0.05in}
\end{figure}

What is of
note here is that the output from every crossbar column in every iteration
is being resolved with a high-precision 9-bit ADC, but many of these bits
contribute to either the 10 least significant bits or the 13 most significant
bits that are eventually going to be ignored.  This is an opportunity to lower
the ADC precision and ignore some bits, depending on the column and the
iteration being processed.  Figure~\ref{adcgrid} shows the number of relevant
bits emerging from every column in every iteration. Note that before dropping 
the highest ignored least significant bit, we use rounding modes to 
generate carries, similar to ~\cite{guptaagrawal15}.

The ADC accounts for a significant fraction of IMA power.  When the ADC is
operating at a lower resolution, it has less work to do.
In every 100~ns iteration, we tune the resolution of a SAR
ADC to match the requirement in Figure~\ref{adcgrid}.
Thus, the use of adaptive ADCs helps reduce IMA power while having no
impact on performance.  We are also ignoring bits that do not show up
in a 16-bit fixed-point result, so we are not impacting the functional
behavior of the algorithm, thus having zero impact on algorithm accuracy.

A SAR ADC does a binary search over the input voltage to find the digital
value,
starting from the MSB.  A bit is set to 1, and the resulting digital value is
converted to analog and compared with the input voltage.  If the input voltage
is higher, the bit is set to one, the next bit is changed, and the process
repeats.
If the number of bits to be sampled is reduced, the circuit can ignore the
latter stages.  The ADC simply gates off its circuits until the next sample
is provided. 
It is important to note that the ADC starts the binary search from the MSB,
thus it is
not possible to sample just the lower significant bits of an output without
knowing
the MSBs. But in this case, we have a unique advantage: if any of the MSBs to
be
truncated is 1, then the output neuron value is clamped to the
highest value in the fixed point range. Thus, in order to sample a set of LSBs,
the ADC starts the binary search with the LSB+1 bit. If that comparison yields
true, it means at least one of the MSB bits is one. 
This signal is sent across the inter-crossbar network (e.g. HTree) and the output is clamped.

In conventional SAR ADCs~\cite{vermachandrakasan07}, a third of the power is
dissipated in the capacitive DAC (CDAC), a third in digital circuits, and a
third in other analog circuits.
The MSB decision in general consumes more power because it involves
charging up the CDAC at the end of every sampling iteration. Recent trends
show CDAC power diminishing due to use of tiny unit capacitances (about 2fF)
and innovative reference buffer designs, leading to ADCs consuming more
power in analog and digital circuits~\cite{kulltoifl13,murmann15}.
The Adaptive ADC technique is able to
reduce energy consumption irrespective of the ADC design since it eliminates
both LSB and MSB tests across the 16 iterations.

\subsection{Communication and Storage Optimizations}

So far, we have discussed optimization techniques that are applicable to any analog
crossbar architectures. To further improve analog accelerator efficiency, it is
critical to also reduce communication and storage overhead. As
the effectiveness of optimizing communication varies based on the overall
architecture, we describe our proposals in the context of the ISAAC architecture. 
Similar to ISAAC, we employ a tiled architecture, where every tile
is composed of several IMAs. A set of IMAs along with digital computational
units and eDRAM storage form a tile. 
%In this context, the three innovations discussed earlier are applied at IMA
%level and the following innovations are applied at tile level. 
%
%\subsection{Intra-Tile Optimizations}
%\label{intratile}
%\vspace{-0.05in}

The previous sub-section focused on techniques to improve an IMA; we
now shift our focus to the design of a tile.  We first reduce the size
of the buffer in each tile that feeds all its IMAs.  We then create
heterogeneous tiles that suit convolutional and fully-connected layers.

\subsubsection{Reducing Buffer Sizes.}
\label{buffer}

\begin{figure*}[th!]
\centering
\includegraphics[width=\textwidth]{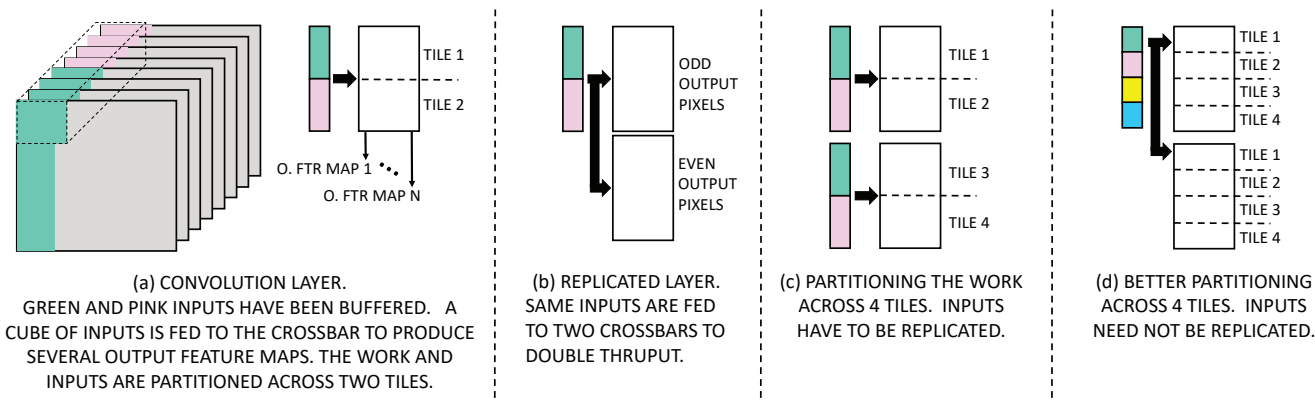}
\vspace{-0.5in}
\caption{Mapping of convolutional layers to tiles.}
\label{fig:buffer}
\vspace{-0.15in}
\end{figure*}

Because ISAAC did not place constraints on how layers are mapped to 
crossbars and tiles, the eDRAM buffer was sized to 64KB to accommodate
the worst-case requirements of workloads.  Here, we design mapping
techniques that reduce storage requirements per tile and move that
requirement closer to the average-case.

To explain the impact of mapping on buffering requirements, first consider
the convolutional layer shown in Figure~\ref{fig:buffer}a.  Once a certain
number of inputs are buffered (shown in green and pink), the layer enters
steady state; every new input pixel allows the convolution to advance by
another step.  The buffer size is a constant as the convolution advances
(each new input evicts an old input that is no longer required).  In
every step, a subset of the input buffer is fed as input to the crossbar
to produce one pixel in each of many output feature maps.  If the crossbar
is large, it is split across 2 tiles, as shown in Figure~\ref{fig:buffer}a.
The split is done so that Tile 1 manages the green buffer and green inputs,
and Tile 2 manages the pink buffer and pink inputs.  Such a split means
that inputs do not have to be replicated on both tiles, and buffering
requirements are low.

Now, consider an early convolutional layer.  Early convolutional layers have
more work to do than later layers since they deal with larger feature maps.
In ISAAC, to make the pipeline balanced, early convolutional layers are
replicated so their throughput matches those of later layers.  
Figure~\ref{fig:buffer}b replicates the crossbar; one is responsible for
every odd pixel in the output feature maps, while the other is responsible
for every even pixel.  In any step, both crossbars receive very similar
inputs.  So the same input buffer can feed both crossbars.

If a replicated layer is large enough that it must be spread across (say) 4 tiles,
we have two options.  Figure~\ref{fig:buffer} c and d show these two options.
If the odd computation is spread across two tiles (1 and 2) and the even
computation is spread across two different tiles (3 and 4), the same green
inputs have to be sent to Tile 1 and Tile 3, i.e., the input buffers are
replicated.  Instead, as shown in Figure~\ref{fig:buffer}d, if we co-locate
the top quadrant of the odd computation and the top quadrant of the even
computation in Tile 1, the green inputs are consumed entirely within Tile 1
and do not have to be replicated.  This partitioning leads to the minimum
buffer requirement.

\begin{figure}[h!]
\centering
\includegraphics[width=\columnwidth]{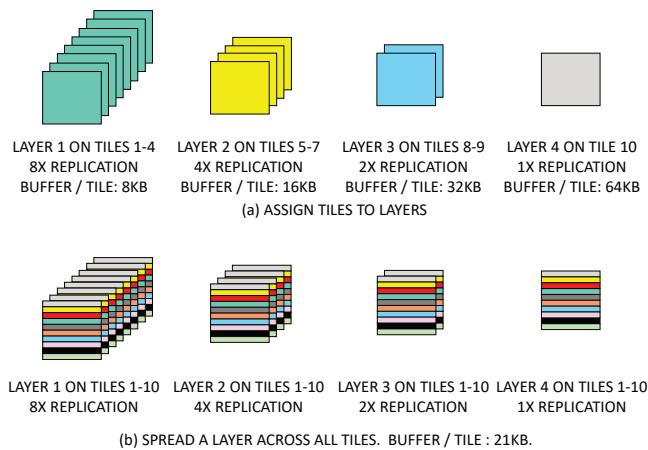}
\vspace{-0.3in}
\caption{Mapping layers to tiles for small buffer sizes.}
\label{fig:avg}
\vspace{-0.15in}
\end{figure}

{\em The bottomline from this mapping is that when a layer is replicated, the
buffering requirements per neuron and per tile are reduced}.  This is because
multiple neurons that receive similar inputs can reuse the contents of the input
buffer.  Therefore, heavily replicated (early) layers have lower buffer
requirements {\em per tile} than lightly replicated (later) layers.
If we mapped these layers to tiles as shown in Figure~\ref{fig:avg}a,
the worst-case buffering requirement goes up (64~KB for the last layer),
and early layers end up under-utilizing their 64~KB buffer.  To reduce
the worst-case requirement and the under-utilization, we instead map
layers to tiles as shown in Figure~\ref{fig:avg}b.  Every
layer is finely partitioned and spread across 10 tiles, and every tile
maps part of a layer.  {\em By spreading each layer across many tiles, every
tile can enjoy the buffering efficiency of early layers.}
By moving every tile's buffer requirement
closer to the average-case (21~KB in this example), we can design a
tile with a smaller eDRAM buffer (21~KB instead of 64~KB) that
achieves higher overall computational efficiency.  This has minimal impact on  
inter-tile neuron communication because adjacent layers are mapped to the same 
tile and hence, even though a single layer is distributed across multiple tiles, 
the neurons being communicated across layers have to typically travel short distances.

\subsubsection{Different Tiles for Convolutions and Classifiers.}
\label{fcconv}

While ISAAC uses the same homogeneous tile for the entire chip, we observe
that convolutional layers have very different resource demands than
fully-connected classifier layers.  The classifier (or FC) layer has to aggregate
a set of inputs required by a set of crossbars; the crossbars then perform
their computation; the inputs are discarded and a new set of inputs is
aggregated.  This results in the following properties for the classifier
layer:
\begin{enumerate}
\item The classifier layer has a high communication-to-compute ratio, so
the router bandwidth puts a limit on how often the crossbars can be busy.
\item The classifier also has the highest synaptic weight requirement 
because every neuron has private weights.
\item The classifier has low buffering requirements -- an input is seen
by several neurons in parallel, and the input can be discarded right after.
\end{enumerate}

We therefore design special tiles customized for classifier layers that:

\begin{enumerate}
\item have a higher crossbar-to-ADC ratio (4:1 instead of 1:1),
\item operate the ADC at a lower rate (10~Msamples/sec instead of 1.2~Gsamples/sec),
\item have a smaller eDRAM buffer size (4~KB instead of 16~KB).
\end{enumerate}

For small-scale workloads that are trying to fit on a single chip, we
would design a chip where many of the tiles are conv-tiles and
some are classifier-tiles (a ratio of 1:1 is a good fit for most of our
workloads).  For large-scale workloads that use multiple chips, each
chip can be homogeneous; we use roughly an equal number of conv-chips and
classifier-chips.  The results consider both cases.

\subsection {Putting the Pieces Together}
\label{implementation}
%\noindent {\bf Implementation}
We use ISAAC as the baseline architecture and evaluate the proposed
optimizations by enhancing it. We already presented a general overview of
ISAAC in section~\ref{isaac}. We make two key enhancements to ISAAC
to improve both area and compute efficiencies. Note that these two
optimizations are specific to
ISAAC architecture, and following this, we present implementation details of numerical
algorithms discussed in previous sub-sections.

First, ISAAC did not place any constraints on how a neural network can be mapped
to its many tiles and IMAs.  As a result, its resources, notably the 
HTree and buffers within an IMA, are provisioned to handle the worst case.
This has a negative impact on power and area efficiency.  Instead, we place
constraints on how the workload is mapped to IMAs.  While this inflexibility
can waste a few resources, we observe that it also significantly reduces
the HTree size and hence area per IMA.
The architecture is still general-purpose, i.e., arbitrary CNNs can be
mapped to it.

Second, 
within an IMA, we co-locate an ADC with each crossbar.
The digitized outputs are then
sent to the IMA's output register via an HTree network.  While ISAAC was
agnostic to how a single synaptic weight was scattered across multiple
bitlines, we adopt the following approach to boost efficiency.  A
16-bit weight is scattered across 8 2-bit cells; each cell is placed in 
a different crossbar.  Therefore, crossbars 0 and 8 are responsible for the
least significant bits of every weight, and crossbars 7 and 15 are responsible
for the most significant bits of every weight.  We also embed the shift-and-add
units in the HTree.  So the shift-and-add unit at the leaf of the HTree
adds the digitized 9-bit dot-product results emerging from two neighboring 
crossbars.  Because the operation is a shift-and-add, it produces a 11-bit
result.  The next shift-and-add unit takes 2 11-bit inputs to produce
a 13-bit input, and so on.   
We further modify mapping by placing the constraint that an IMA
cannot be shared by multiple network layers.

\begin{figure}[h!]
        \centering
        \includegraphics[scale=1.2]{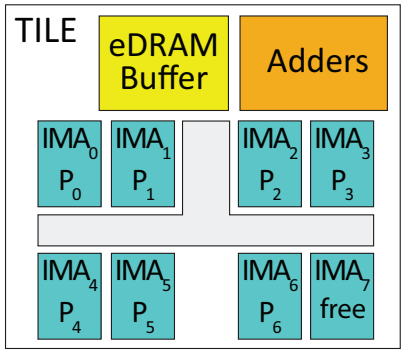}
        %\vspace{-0.1in}
        \caption{Mapping Strassen's algorithm to a tile.}
        \label{fig:strassentile}
        \vspace{-0.1in}
\end{figure}

\begin{figure}
\centering
\includegraphics[width=\columnwidth]{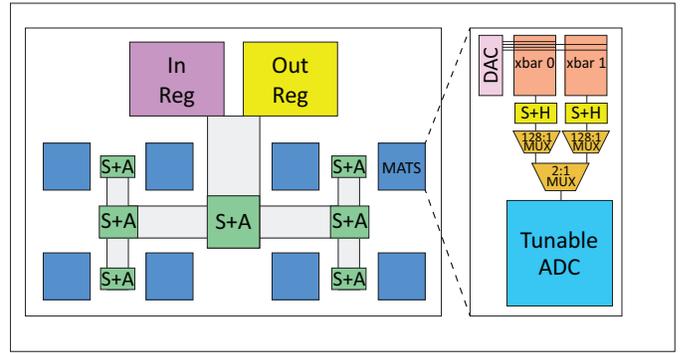}
\vspace{-0.15in}
\caption{IMA supporting Karatsuba's Algorithm.}
\label{fig:newima}
\vspace{-0.2in}
\end{figure}

To implement Karatsuba's algorithm, we modify the
In-situ Multiply Accumulate units (IMA) as shown in Figure~\ref{fig:newima}.
The changes are localized to a single mat.  Each mat now has two crossbars that
share the DAC and ADC.  Given the size of the ADC, the extra crossbar per mat
has a minimal impact on area.  The left crossbars in four of the mats now store
$W_0$ (Figure~\ref{fig:karatsuba}); the left crossbars in the other four mats
store $W_1$; the right crossbars in five of the mats store $W_0+W_1$; the right
crossbars in three of the mats are unused.  In the first 8 iterations, the 8
ADCs are used by the left crossbars.  In the next 9 iterations, 5 ADCs are used
by the right crossbars.  As discussed earlier, the main objective here is to
lower power by reducing use of the ADC.  Divide \& Conquer can be recursively
applied further.  When applied again, the computation keeps 8 ADCs busy in the
first 4 iterations, and 6 ADCs in the next 10 iterations.  This is a 28\%
reduction in ADC use, and a 13\% reduction in execution time.  But, we pay an
area penalty because 20 crossbars are needed per IMA.
Figure~\ref{fig:strassentile} shows the mapping of computations within IMA to
implement Strassen's algorithm. The computations ($P_0-P_6$) in Strassen's
algorithm (Figure~\ref{fig:strassen}) are mapped to 7 IMAs in the tile.  The
8th IMA can be allocated to another layer's computation.

With all these changes targeting high compute efficiency and low communication
and storage overhead, we refer to the updated analog design as the \name
architecture.

%\begin{figure}
%\centering
%\includegraphics[width=\columnwidth]{figures/htreemodifications2.eps}
%\vspace{-0.15in}
%\caption{Microarchitecture of an IMA.}
%\label{fig:oldima}
%\vspace{-0.05in}
%\end{figure}

\vspace{-0.00in}
\section{Methodology}
\label{method}
\vspace{-0.0in}

\noindent {\bf Modeling Area and Energy}

For modeling the energy and area of the eDRAM buffers and on-chip 
interconnect like the HTree and tile bus, we use CACTI 6.5~\cite{muralimanoharbalasubramonian07b}
at 32~nm. The area and energy model of a memristor crossbar is based on \cite{hustrachan16}.
We adapt the area and energy of shift-and-add circuits, max/average pooling block and 
sigmoid operation similar to the analysis in DaDianNao~\cite{chenluo14} and
ISAAC~\cite{shafieenag16}. We avail the same HyperTransport serial link model for off-chip 
interconnects as used by DaDianNao~\cite{chenluo14} and ISAAC~\cite{chenluo14}. The 
router area and energy is modeled using Orion 2.0~\cite{kahngli09}.
While our buffers can also be implemented with SRAM, we use eDRAM to make an
apples-to-apples comparison with the ISAAC baseline.
\name is only used for inference, with a delay of 16.4~ms to pre-load weights
in a chip.

In order to model the ADC energy and area, we use a recent survey~\cite{murmann15} of 
ADC circuits published in different circuit conferences. The \name architecture uses 
the same 8-bit ADC~\cite{kulltoifl13} at 32~nm as used in ISAAC, partly because it yields the best 
configuration in terms of area/power and meets the sampling frequency
requirement, and 
partly because it can be reconfigured for different resolutions. This is at 
the cost of minimal increase in area of the ADC. We scale the ADC power with respect to 
different sampling frequency according to another work by Kull et al.~\cite{kulltoifl13}.
The SAR ADC has six different components: comparators, asynchronous clock logic,
sampling clock logic, data memory and state logic,
reference buffer, and capacitive DAC. The ADC power for different sampling resolution 
is modeled by gating off the other components except the sampling clock.

We consider a 1-bit DAC as used in ISAAC because it is relatively small and has 
high SNR value. Since DAC is used in every row of the crossbar, a 1-bit DAC improves 
the area efficiency. 

The key parameters in the architecture that 
largely contribute 
to our analysis are reported in Table~\ref{newtonparams}.

This work considers recent workloads with state-of-the-art accuracy in
image classification tasks (summarized in Table~\ref{bench}). We create an analytic model for a \name pipeline 
within an IMA and within a tile and map the suite of benchmarks, making sure that 
there are no structural hazards in any of these pipelines. 
We consider network bandwidth limitations in our simulation model to
estimate throughput.
Since ISAAC is a throughput architecture, we do an iso-throughput comparison
of the \name architecture with ISAAC for the different intra-IMA or 
intra-tile optimizations.
Since the dataflow in the architecture is bounded by the router bandwidth,
in each case, we allocate enough resources till the network saturates to
create our baseline model.  For subsequent optimizations, we retain the
same throughput.
%The pipelines work 
%at a granularity of between layers and within a layer of a deep neural network.
Similar to ISAAC, data transfers between tiles on-chip, and on the HT link across 
chips have been statically routed to make it conflict free. Like ISAAC, the latency 
and throughput of \name for the given benchmarks can be calculated analytically 
using a deterministic execution model. Since there aren't any run-time dependencies 
on the control flow or data flow of the deep networks, analytical estimates 
are enough to capture the behavior of cycle-accurate simulations.

We create a similar model for ISAAC, taking into considerations all the parameters 
mentioned in their paper.

\noindent {\bf Design Points}

The \name architecture can be designed by optimizing one of the following two metrics:
\begin{enumerate}
\item {\bf CE:} Computational Efficiency which is the number of fixed point operations 
performed per second per unit area, $GOPS/(s\times mm^2)$.
\item {\bf PE:} Power Efficiency which is the number of fixed point operations
performed per second per unit power, $GOPS/(s\times W)$.
\end{enumerate}

For every considered innovation, we model \name for a variety of design points that
vary crossbar size, number of crossbars per IMA, and number of IMAs per tile.  In
most cases, the same configurations emerged as the best.  We therefore focus most
of our analysis on this optimal configuration that has 16 IMAs per tile, where
each IMA uses 16 crossbars to process 128 inputs for 256 neurons.
We report the area, power, and 
energy improvement for all the deep neural networks in our benchmark suite.

\begin{table}[h!]
\footnotesize
\centering
\begin{tabular}{| c | c | c | c |} 
\hline
{\bf Component} & {\bf Spec} & {\bf Power} & {\bf Area} ($mm^2$) \\
\hline
Router & 32 flits, 8 ports & 168 mW & 0.604 \\
\hline
ADC & 8-bit resolution & 3.1 mW &  0.0015   \\
	& 1.2 GSps frequency & & \\
\hline
Hyper Tr & 4 links @ 1.6GHz & 10.4 W & 22.88 \\
 & 6.4 GB/s link bw & & \\
\hline
DAC array & 128 1-bit resolution  & 0.5 mW & 0.00002 \\
	& number & $8\times 128$ & \\
\hline
Memristor crossbar & $128 \times 128$ & 0.3 mW & 0.0001 \\
\hline
 \end{tabular}
 \vspace{0.1in}
 \caption{Key contributing elements in \names.}
 \label{newtonparams}
\vspace{-0.2in}
\end{table}

\begin{table*}[ht]
\footnotesize
\centering
 \begin{tabular}{|c | c | c | c | c | c | c | c| c | c |} 
 \hline
\textbf{input}  & \textbf{Alexnet} & \textbf{VGG-A}& \textbf{VGG-B} & \textbf{VGG-C} & \textbf{VGG-D} &\textbf{MSRA-A} & \textbf{MSRA-B} & \textbf{MSRA-C} & \textbf{Resnet-34}\\ [0.5ex] 
\textbf{size} & \cite{krizhevskysutskever12} & \cite{simonyanzisserman14} & \cite{simonyanzisserman14} & \cite{simonyanzisserman14} & \cite{simonyanzisserman14} & \cite{hezhang15} & \cite{hezhang15} & \cite{hezhang15} & \cite{hezhang15b}\\
 \hline\hline
224 & 11x11, 96 (4) & 3x3,64 (1) & 3x3,64 (2) & 3x3,64 (2) & 3x3,64 (2) & 7x7,96/2(1) & 7x7,96/2(1) & 7x7,96/2(1) & 7x7,64/2\\
\hline
& 3x3 pool/2 & \multicolumn{4}{c|}{2x2 pool/2}& \multicolumn{3}{c |}{} & 3x3 pool/2\\
\hline
112 & & 3x3,128 (1) & 3x3,128 (2) & 3x3,128 (2) & 3x3,128 (2) & & & & \\
\hline
& & \multicolumn{7}{c|}{2x2 pool/2} & \\
 \hline
56 & & 3x3,256 (2) & 3x3,256 (2) & 3x3,256 (3) & 3x3,256 (4) & 3x3,256 (5) & 3x3,256 (6) & 3x3,384 (6) & 3x3,64 (6)\\
 & & & 1x1, 256(1) & & & & & &\\
\hline
& &\multicolumn{7}{c|}{2x2 pool/2} & 3x3,128/2(1)\\
 \hline
28 & 5x5,256 (1) & 3x3,512 (2) & 3x3,512 (2) & 3x3,512 (3) & 3x3,512 (4) & 3x3,512 (5) & 3x3,512 (6) & 3x3,768 (6) & 3x3,128 (7)\\
   &             &             & 1x1,256 (1) &             &             &             &             &             &\\
\hline
& 3x3 pool/2 & \multicolumn{7}{c|}{2x2 pool/2} & 3x3,256/2 (1)\\
 \hline
14 & 3x3,384 (2) & 3x3,512 (2) & 3x3,512 (2) & 3x3,512 (3) & 3x3,512 (4) & 3x3,512 (5) & 3x3,512 (6) & 3x3,896 (6) & 3x3,256 (11)\\
   & 3x3,256 (1) &             & 1x1,512 (1) &             &             &             &             &             & \\
\hline
   & 3x3 pool/2  &\multicolumn{4}{c|}{2x2 pool/2}                        & \multicolumn{3}{c |}{spp,{7,3,2,1}}     & 3x3,512/2 (1)\\
 \hline 
7 &\multicolumn{8}{c|}{FC-4096(2)} & 3x3,512 (5)\\
 \hline 
 &\multicolumn{9}{c|}{FC-1000(1)}\\
 \hline  
 \end{tabular}
 \vspace{0.1in}
 \caption{Benchmark names are in bold. Layers are formatted as $K_x\times K_y, N_o$/stride (t), where t is the number of such layers. Stride is 1 unless explicitly mentioned. Layer* denotes convolution layer with private kernels.}
 \label{bench}
\vspace{-0.25in}
\end{table*}

\vspace{-0.10in}
\section{Results}
\label{results}

%As we move through the results, we will incrementally add each innovation.
%Each reported improvement is relative to the \name architecture with
%innovations described until that point.
The \name architecture takes the baseline analog accelerator ISAAC and incrementally applies 
innovations discussed earlier.
We begin by describing results for optimizations targeting global components
such as h-tree followed by tile and IMA
level techniques. As mentioned earlier, while we build on
ISAAC and use it for evaluation, the proposed enhancements to
crossbar  are applicable to any analog architecture.

\noindent {\bf Constrained Mapping for Compact HTree}

\begin{figure}[h!]
\centering
\includegraphics[width=\columnwidth]{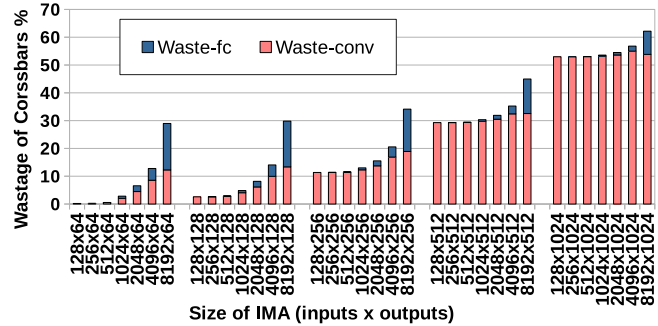}
\vspace{-0.20in}
\caption{Xbar under-utilization with constrained mapping.}
\label{wastage}
\vspace{-0.2in}
\end{figure}

\begin{figure}[h!]
\centering
\includegraphics[width=\columnwidth]{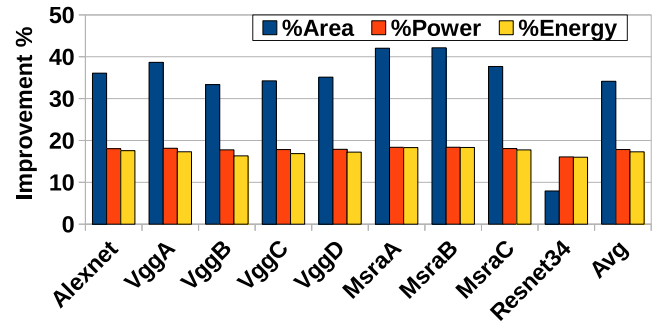}
\vspace{-0.25in}
\caption{Impact of constrained mapping and compact HTree.}
\label{htreemods}
\vspace{-0.05in}
\end{figure}

We first observe that the ISAAC IMA is designed with an 
over-provisioned HTree that can handle a worst-case mapping of the workload. 
We imposed the constraint that an IMA can only handle a single layer, and a maximum of
128 inputs.  This restricts the width of the HTree, promotes input sharing, and enables
reduction of partial neuron values at the junctions of the HTree.  While this helps
shrink the size of an IMA, it suffers from crossbar under-utilization within an IMA.
We consider different IMA sizes, ranging 
from $128\times64$ which supplies the same 128 neurons to 4 crossbars to get 64 output neurons, to 
$8192\times1024$. 
Figure~\ref{wastage} plots the average under-utilization of crossbars across the different 
workloads in the benchmark suite.  For larger IMA sizes, the under-utilization is quite
significant.  Larger IMA sizes also result in complex HTrees.  Therefore, a moderately
sized IMA that processes 128 inputs for 256 neurons has high computational efficiency and
low crossbar under-utilization.  For this design, the under-utilization is only 9\%.
Figure~\ref{htreemods} quantifies how our constrained mapping and compact
HTree improve area, power, and energy per workload.  In short, our
constraints have improved area efficiency by 37\% and power/energy
efficiency by 18\%, while leaving only 9\% of crossbars under-utilized.

\noindent {\bf Heterogeneous ADC Sampling}

\begin{figure}[h!]
\centering
\includegraphics[width=\columnwidth]{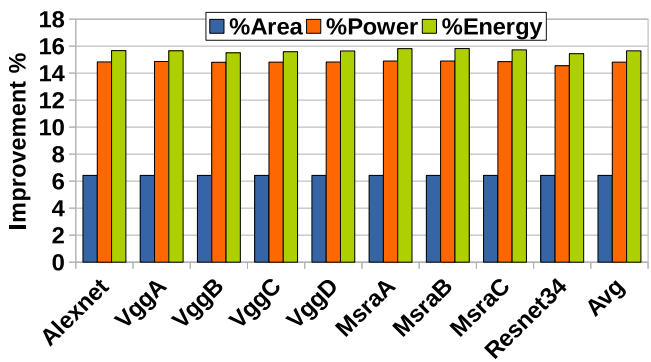}
\vspace{-0.30in}
\caption{Improvement due to the adaptive ADC scheme.}
\label{adaptiveadcimprovement}
\vspace{-0.15in}
\end{figure}

The heteregenous sampling of outputs using adaptive ADCs has a big impact on reducing the power profile 
of the analog accelerator. In one iteration of 100 ns, at max 4 ADCs work at the max resolution of 8-bits. Power 
supply to the rest of the ADCs can be reduced. We measure the reduction of area, power, and energy with respect to 
the new IMA design with the compact HTree.
Since ADC contributed to 49\% of the chip power in ISAAC, 
reducing the oversampling of ADC reduces power requirement by 15\% on average. The area efficiency improves as well 
since the output-HTree now carries 16-bits instead of unnecessarily carrying 39-bits of final output. The improvements 
are shown in Figure~\ref{adaptiveadcimprovement}.
%\vspace{-0.05in}

\noindent {\bf Karatsuba's Algorithm}

\begin{figure}[h!]
\centering
\includegraphics[scale=0.5]{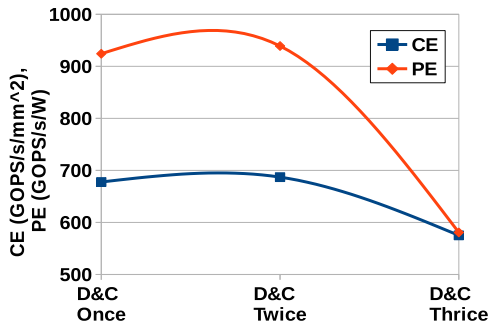}
\vspace{-0.05in}
\caption{Comparison of CE and PE for Divide and Conquer done recursively.}
\label{dccomparison}
\vspace{-0.05in}
\end{figure}

\begin{figure}[h!]
\centering
\includegraphics[width=\columnwidth]{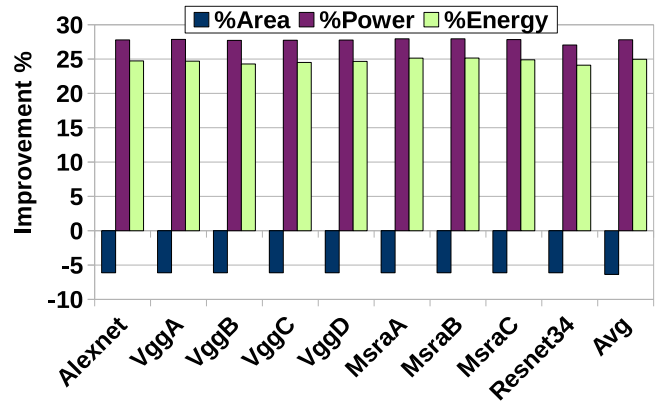}
\vspace{-0.25in}
\caption{Improvement with Karatsuba's Algorithm.}
\label{dcimprovement}
\vspace{-0.15in}
\end{figure}

We further try to reduce the power profile with divide-and-conquer within
an IMA.
Figure ~\ref{dccomparison} shows the impact of recursively applying the
divide-and-conquer technique multiple times.  Applying it once is nearly as 
good as applying it twice, and much less complex.  Therefore, we focus
on a single divide-and-conquer step.
Improvements are reported in Figure~\ref{dcimprovement}. Energy efficiency 
improves by almost 25\% over the previous design point, because ADCs end up being used 75\% of the times in the 
1700 ns window. However, this comes at a cost of 6.4\% reduction in area efficiency because of the need for more 
crossbars and increase in HTree bandwidth to send the sum of inputs.

\noindent {\bf eDRAM Buffer Requirements}

\begin{figure}[h!]
\centering
\includegraphics[scale=0.6]{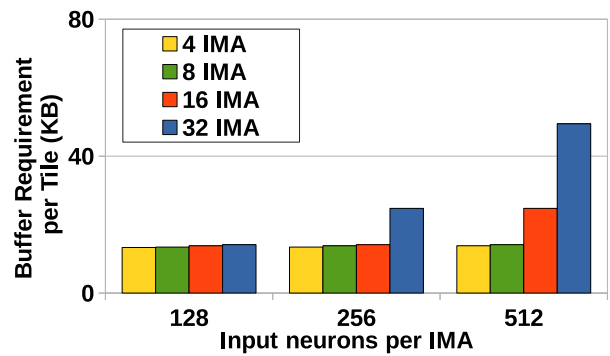}
\vspace{-0.05in}
\caption{Buffer requirements for different tiles, changing the type of IMA and the number of IMAs.}
\label{bufferrequirement}
\vspace{-0.05in}
\end{figure}

\begin{figure}[h!]
\centering
\includegraphics[width=\columnwidth]{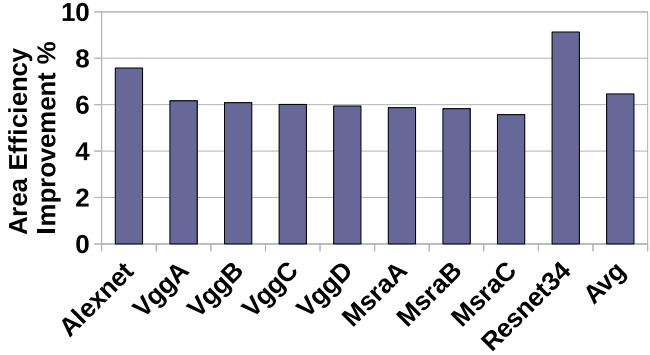}
\vspace{-0.25in}
\caption{Improvement in area efficiency with decreased eDRAM buffer sizes.}
\label{bufferreduction}
\vspace{-0.2in}
\end{figure}

In Figure~\ref{bufferrequirement}, we report the buffer requirement per tile when the layers are spread across many
tiles.  We consider this for a variety of tile/IMA configurations.
Image size has a linear impact on the buffering requirement. For 256$\times$256 images, 
the buffer reduction technique leads to the choice of a 16 KB buffer instead of the 64 KB 
used in ISAAC, a 75\% reduction. Figure~\ref{bufferreduction} shows 6.5\% average 
improvement in area efficiency because of this technique.

\noindent {\bf Conv-Tiles and Classifier-Tiles}

\begin{figure}[h!]
\centering
\includegraphics[width=\columnwidth]{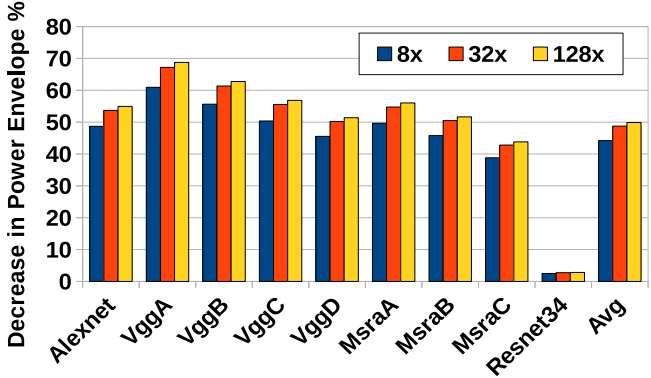}
\vspace{-0.25in}
\caption{Decrease in power requirement when frequency of FC tiles is altered.}
\label{frequencysweep}
\vspace{-0.2in}
\end{figure}

\begin{figure}[h!]
\centering
\includegraphics[width=\columnwidth]{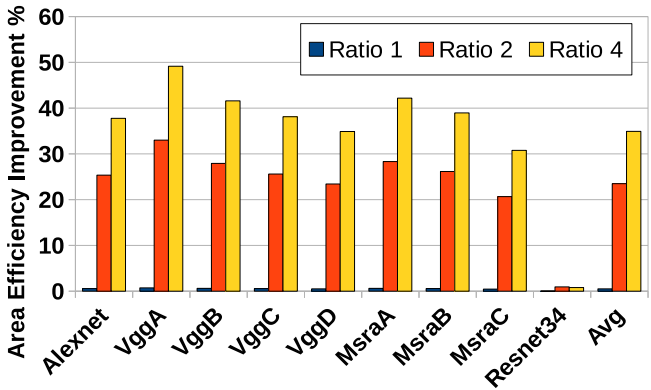}
\vspace{-0.25in}
\caption{Improvement in area efficiency when sharing multiple crossbars per ADC in FC tiles.}
\label{ratiosweep}
\vspace{-0.05in}
\end{figure}

Figure~\ref{frequencysweep} plots the decrease in power requirement when FC tiles are operated at $8\times$, $32\times$ and 
$128\times$ slower than the conv tiles. None of these configurations lower the throughput as the FC layer
is not on the critical path.
Since ADC power scales linearly with sampling resolution, 
the power profile is lowest when the ADCs work $128\times$ slower. This leads to 50\% lower peak power on average.
In Figure~\ref{ratiosweep}, we plot the increase in area efficiency when multiple crossbars share the same ADC in FC tiles. 
The underutilization of FC tiles provides room for making them storage efficient, saving on average 38\% 
of chip area. We do not increase the ratio beyond 4 because the multiplexer connecting the crossbars to the ADC becomes complex.
Resnet does not gain much from the heterogeneous tiles because it needs relatively fewer FC tiles.

\vspace{-0.05in}

\noindent {\bf Strassen's Algorithm}

\begin{figure}[h!]
\centering
\includegraphics[width=\columnwidth]{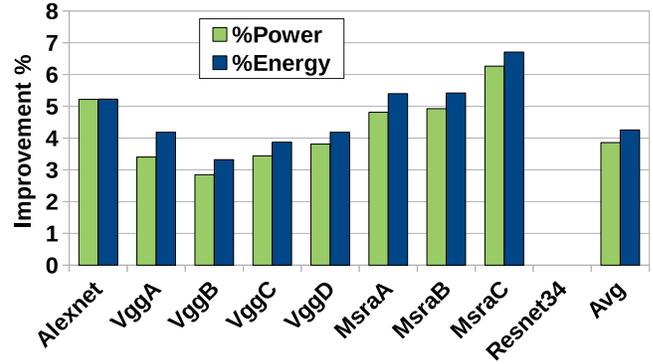}
\vspace{-0.25in}
\caption{Improvement due to the Strassen technique.}
\label{strassenimprovement}
\vspace{-0.05in}
\end{figure}

Strassen's optimization is especially useful when large matrix multiplication can be performed in the conv layers 
without much wastage of crossbars. This provides room for decomposition of these large matrices, which is the key part 
of Strassen's technique. We note that Resnet has high wastage when using larger IMAs, and thus does not benefit at all from 
this technique. Overall, Strassen's algorithm increases the energy efficiency by 4.5\% as seen in Figure~\ref{strassenimprovement}. 

\vspace{-0.05in}

\noindent {\bf Putting it all together.}

\begin{figure}[h!]
\centering
\includegraphics[width=\columnwidth]{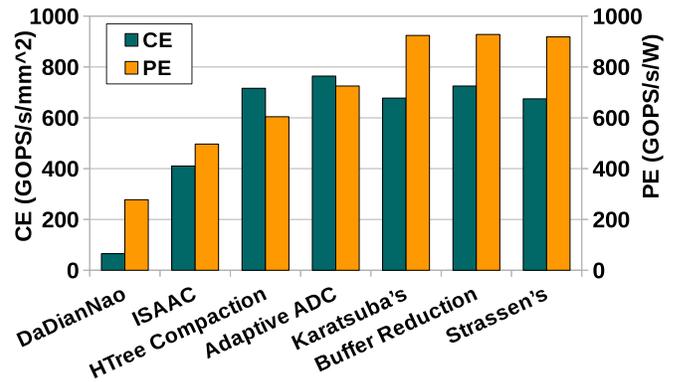}
\vspace{-0.25in}
\caption{Peak CE and PE metrics of different schemes along with baseline digital and analog accelerator.}
\label{ceandpe}
\vspace{-0.10in}
\end{figure}

Figure~\ref{ceandpe} plots the incremental effect of each of our techniques on peak computational and
power efficiency of DaDianNao, ISAAC, and \names. 
We do not include the heterogeneous FC tile in this plot because it is forcibly operated slowly because it
is non-critical; as a result, it's peak throughput is lower by definition.
We see that both adaptive ADC and divide \& conquer play a significant role in increasing the PE. 
While the impact of Strassen's technique is not visible in this graph, it manages to free up resources (1 every 8 IMA) in a tile, thus 
providing room for more compact mapping of networks, and reducing ADC utilization.

\begin{figure}[h!]
\centering
\includegraphics[width=\columnwidth]{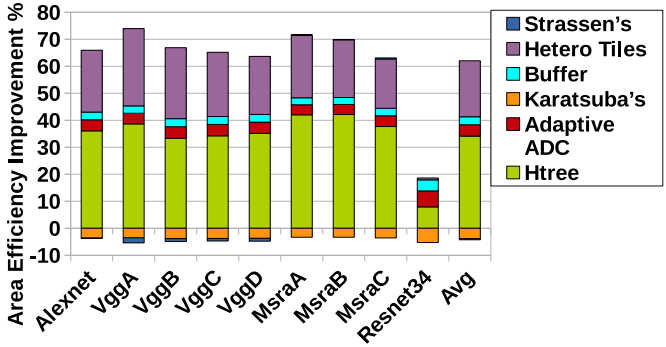}
\vspace{-0.25in}
\caption{Breakdown of area efficiency.}
\label{areaimprovement}
\vspace{-0.25in}
\end{figure}

\begin{figure}[h!]
\centering
\includegraphics[width=\columnwidth]{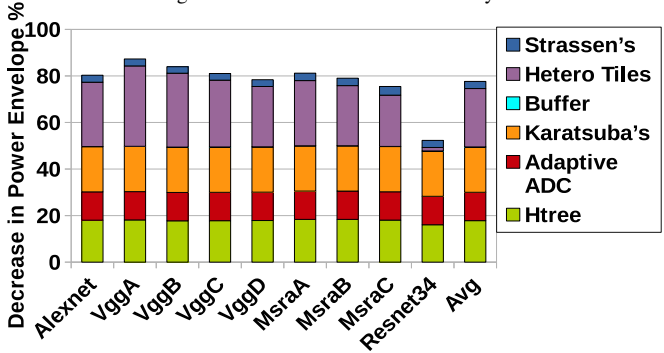}
\vspace{-0.25in}
\caption{Breakdown of decrease in power envelope.}
\label{powerimprovement}
\vspace{-0.25in}
\end{figure}

\begin{figure}[h!]
\centering
\includegraphics[width=\columnwidth]{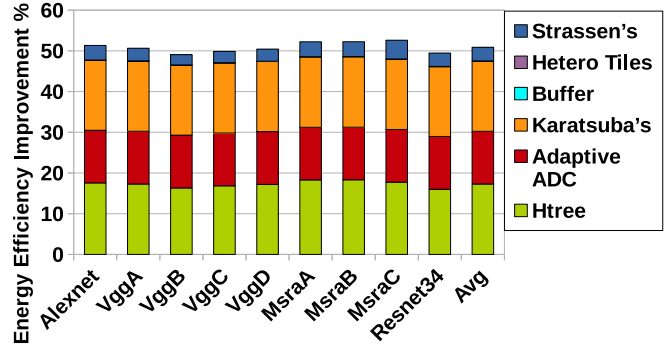}
\vspace{-0.25in}
\caption{Breakdown of energy efficiency.}
\label{energyimprovement}
\vspace{-0.05in}
\end{figure}

Figure~\ref{areaimprovement} shows a per-benchmark improvement in area efficiency and the contribution 
of each of our techniques.  The compact HTree and the FC tiles are the biggest contributors.
Figure~\ref{powerimprovement} similarly shows a breakdown for 
decrease in power envelope, and Figure~\ref{energyimprovement} does the same of improvement in energy efficiency.
Multiple innovations (HTree, adaptive ADC, Karatsuba, FC tiles) contribute equally to the improvements.
We also observed that the Adaptive ADC technique's improvement is not very
sensitive to the ADC design.  We evaluated ADCs where the CDAC power dissipates
10\% and 27\% of ADC power; the corresponding improvements with the Adaptive
ADC were 13\% and 12\% respectively.

\begin{figure}[h!]
\centering
\includegraphics[width=\columnwidth]{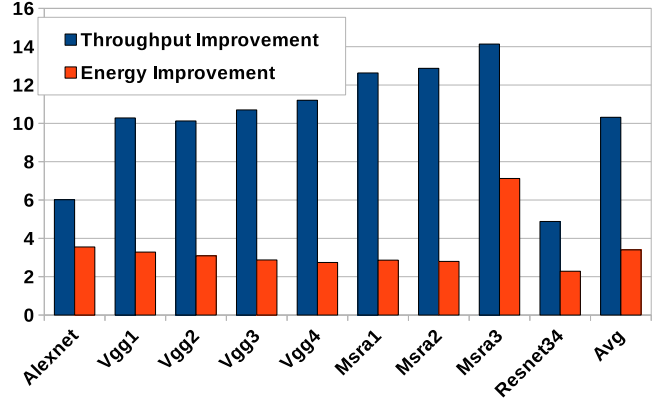}
\vspace{-0.25in}
\caption{Comparison with TPU}
\label{tpucomparison}
\vspace{-0.05in}
\end{figure}

Figure~\ref{tpucomparison} compares the 8-bit version of \names with Google's
TPU architecture. Note that while Google has already announced second
generation TPU with 16-bit support, its architectural details are not public
yet. Hence, we limit our analysis to TPU-1. Also, we scale
the area such that the die area is same for both the architectures, i.e. an iso-area comparison.
%
%We also compare the 8-bit version of \names with the latest 8-bit digital accelerator by Google, 
%the TPU. In our analysis, we provide equal resources to each accelerator, i.e. an iso-area 
%comparison. 
For TPU, we perform batch processing enough to not exceed the latency target of 7ms 
as demanded by most application developers. Since \names pipeline  is
deterministic and as its crossbars are statically mapped to different layers,
the latency of images is always the same irrespective of batch size, which is 
comfortably less than 7ms for all the evaluated benchmarks. We also model TPU-1
with GDDR5 memory to allocate sufficient bandwidth. 

Figure~\ref{tpucomparison} shows throughput and energy improvement of \names
over TPU for various benchmarks. 
\names has an average 
improvement of 10.3$\times$ in throughput and 3.4$\times$ in energy over TPU. 
In terms of computational efficiency (CE) calculated using peak throughput,
 \names is 12.3$\times$ better than TPU.
However, when operating on FC layer, due to idle crossbars in
\names, this advantage reduces to 10.3$\times$ for actual workloads.

When considering power efficiency (PE) calculated using peak throughput
and area, although \names is only 1.6$\times$ better than TPU, the actual
benefit goes up for real workloads, increasing it to 3.4$\times$. This is
because of TPU's low  memory bandwidth coupled with reduced batch size for some workloads. 
As we discussed earlier, the batch size in TPU is adjusted to meet the latency
target. Since large batch size alleviates memory bandwidth problem, reducing it
to meet latency target directly impacts power efficiency due to more GDDR
fetches and idle processing units.

%Even though \names is 
%12.3$\times$ better in Computational Efficiency, it wastes some of it in the FC layer, even 
%though not as much as ISAAC since \names uses heteregeneous tiles for CONV and FC layers. TPU on the 
%other hand works way below the peak for FC layers due to bandwidth limitation that can not be 
%amortized using batching if latency targets are to be achieved. \names is 1.6$\times$ better 
%in Power Efficiency but TPU wastes more energy in costly GDDR5 fetches and idle power, 
%because of which \names gets 3.4$\times$ energy improvement. 

From the figure, it can also be noted that the throughput improvement of Alexnet 
and Resnet aren't as high as the other benchmarks because of their relatively
small networks. This increases the batch size, improving the data locality for
FC layer weights.
%which has a high batching factor, thus amortizing the cost of fetching the FC layer weights. 
On the other hand, the 
MSRA3 benchmark has  higher energy consumption than other workloads because for
MSRA3, TPU can process only
one image per batch. This dramatically increases TPU's idle time while fetching a large number 
of weights for the FC layers.
In short, Newton's in-situ computation achieves superior energy and performance
values over TPU as the proposed design limits data movement while reducing analog
computation overhead. 

\vspace{-0.05in}

\vspace{-0.15in}
\section{Conclusions}
\label{conclusions}
\vspace{-0.05in}
In this work, we target resource provisioning and efficiency in a crossbar-based
deep network accelerator.  Starting with the ISAAC architecture, we show
that three approaches -- heterogeneity, mapping constraints, and divide \&
conquer -- can be applied within a tile and within an IMA.  This results
in smaller eDRAM buffers, smaller HTree, energy-efficient ADCs with
varying resolution, energy- and area-efficiency in classifier layers,
and fewer computations.  Many of these ideas would also apply to a
general accelerator for matrix-matrix multiplication, as well as to
other neural networks such as RNN, LSTM, etc.
The \name architecture cuts the current gap between ISAAC and an
ideal neuron in half.

\vspace{0.1in}
\noindent{\bf \large Appendix: Crossbar Implementations}

This appendix discusses how crossbars can be designed to withstand
noise effects in analog circuits.

\noindent {\em Process Variation and Noise:}
Since an analog crossbar uses actual conductance of individual cells to perform
computation, it is critical to do writes at maximum precision. We make two
design choices to improve write precision. First, we equip each cell
with an access transistor (1T1R cell) to precisely control the amount of write
current going through it. While this increases area overhead, it eliminates
sneak currents and their negative impact on write voltage
variation~\cite{zangenehjoshi14}. Second, we use a closed loop write circuit
with current compliance that does many iterations of program-and-verify
operations. Prior work has shown that such an approach can provide more precise
states at the cost of increased write time even with high process variation in
cells~\cite{alibartgao12}. 

%{\em Process Variation and Noise:} To address write
%precision, we use a closed loop write circuit with current
%compliance that does many program-and-verify operations
%to fine tune cells. Prior work has shown that such
%an approach can provide more precise states at the cost
%of increased programming time \cite{alibartgao12}. We further use
%“1T1R” cells that reduce density, but avoid sneak currents
%and their negative impact on write voltage variation
%\cite{zangenehjoshi14}.

In spite of a robust write process, a cell's resistance
will still deviate from its normal value within a tolerable
range. This range will ultimately limit either the
number of levels in a cell or the number of simultaneously
active rows in a crossbar. For example, if a cell write
can achieve a resistance within $\Delta$r ($\Delta$r is a function of
noise and parasitic), if $l$ is the number of levels in a cell,
and $rrange$ is the max range of resistance of a cell,
then we set the number of active rows to $rrange$/($l$.$\Delta$r)
to ensure there are no corrupted bits at the ADC.

\noindent {\em Crossbar Parasitic:} While a sophisticated write circuit coupled with limited
array size can help alleviate process variation and noise, IR drop along rows
and columns can also reduce crossbar accuracy. When a crossbar is being written during
initialization, the access transistors in unselected cells shut off the sneak
current path, limiting the current flow to just the selected cells. However,
when a crossbar operates in compute mode in which multiple rows are active, the net
current in the crossbar increases, and the current path becomes more complicated. With
access transistors in every cell in the selected rows in ON state, a network of
resistors is formed with every cell conducting varying current based on its
resistance. As wire links connecting these cells have non-zero resistance, the
voltage drop along rows and columns will impact the computation accuracy.
Thus, a cell at the far end of the driver will see relatively lower read
voltage compared to a cell closer to the driver. This change in voltage is a
function of both wire resistance and the current flowing through wordlines and
bitlines, which in turn is a function of the data pattern in the array.  This
problem can be addressed by limiting the DAC voltage range and doing data encoding
to compensate for the IR drop~\cite{hustrachan16}. Since the matrix being
programmed into a crossbar is known beforehand, during the initialization phase of a
crossbar, it is possible to account for voltage drops and adjust the cell resistance
appropriately. Hu et al.~\cite{hustrachan16} have demonstrated successful %
%crossbar operations. They showed proper functioning 
operation of a 256$\times$256 crossbar with 5-bit cells
even in the presense of thermal noise in memristor, short noise in circuits,
and random telegraphic noise in the crossbar.
%with better encoding schemes and low range DACs for crossbars of size
%upto 256x256 with 5-bit cells.  
For this work, a conservative model with a 128$\times$128 crossbar
with 2-bit cells and 1-bit DAC emerges as an ideal design point in most
experiments.

\bstctlcite{bstctl:etal, bstctl:nodash, bstctl:simpurl}
\bibliographystyle{IEEEtranS}
\bibliography{main}

\end{document}